\documentclass[pdflatex,sn-mathphys-num]{sn-jnl}

\usepackage{xcolor}
\usepackage{graphicx}%
\usepackage{multirow}%
\usepackage{amsmath,amssymb,amsfonts}%
\usepackage{amsthm}%
\usepackage{mathrsfs}%
\usepackage[title]{appendix}%
\usepackage{xcolor}%
\usepackage{textcomp}%
\usepackage{manyfoot}%
\usepackage{booktabs}%
\usepackage{algorithm}%
\usepackage{algorithmicx}%
\usepackage{algpseudocode}%
\usepackage{listings}%

\theoremstyle{thmstyleone}%
\theoremstyle{thmstyletwo}%

\theoremstyle{thmstylethree}%

\begin{document}

\title[Article Title]{Seasonal Forecasting of Pan-Arctic Sea Ice with State Space Model}


\author[1]{\fnm{Wei} \sur{Wang}}\email{wang\_wei23@m.fudan.edu.cn}

\author*[1]{\fnm{Weidong} \sur{Yang}}\email{wdyang@fudan.edu.cn}

\author[2,3]{\fnm{Lei} \sur{Wang}}\email{wanglei\_ias@fudan.edu.cn}

\author[2]{\fnm{Guihua} \sur{Wang}}\email{wanggh@fudan.edu.cn}

\author[4]{\fnm{Ruibo} \sur{Lei}}\email{leiruibo@pric.org.cn}

\affil*[1]{\orgdiv{School of Computer Science}, \orgname{Fudan University}, \orgaddress{\city{Shanghai}, \country{China}}}

\affil[2]{\orgdiv{Department of Atmospheric and Oceanic Sciences \& Institute of Atmospheric Sciences}, \orgname{Fudan University}, \orgaddress{\city{Shanghai},  \country{China}}}

\affil[3]{\orgdiv{Key Laboratory of Polar Atmosphere-ocean-ice System for Weather and Climate, Ministry of Education}, \orgname{Fudan University}, \orgaddress{ \city{Shanghai},  \country{China}}}

\affil[4]{\orgdiv{Key Laboratory of Polar Science}, \orgname{MNR, Polar Research Institute of China}, \orgaddress{ \city{Shanghai},  \country{China}}}


\abstract{The rapid decline of Arctic sea ice resulting from anthropogenic climate change poses significant risks to indigenous communities, ecosystems, and the global climate system. This situation emphasizes the immediate necessity for precise seasonal sea ice forecasts. While dynamical models perform well for short-term forecasts, they encounter limitations in long-term forecasts and are computationally intensive. Deep learning models, while more computationally efficient, often have difficulty managing seasonal variations and uncertainties when dealing with complex sea ice dynamics. In this research, we introduce IceMamba, a deep learning architecture that integrates sophisticated attention mechanisms within the state space model. Through comparative analysis of 25 renowned forecast models, including dynamical, statistical, and deep learning approaches, our experimental results indicate that IceMamba delivers excellent seasonal forecasting capabilities for Pan-Arctic sea ice concentration. Specifically, IceMamba outperforms all tested models regarding average RMSE and anomaly correlation coefficient (ACC) and ranks second in Integrated Ice Edge Error (IIEE). This innovative approach enhances our ability to foresee and alleviate the effects of sea ice variability, offering essential insights for strategies aimed at climate adaptation.}

\maketitle

\section{Introduction}\label{sec1}

Arctic sea ice plays a critical role in the global climate system, profoundly influencing the atmosphere, oceans, and global weather patterns~\cite{liu2021acceleration, smith2022robust,chripko2021impact}. As global warming intensifies, the retreat of Arctic sea ice has become a notable characteristic of climate change. The decline in sea ice alters the surface albedo, reducing the reflection of solar radiation, leading to increased heat absorption by the exposed ocean, further accelerating the warming in the region~\cite{cai2021accelerated}. This warming can disrupt atmospheric circulation patterns~\cite{chung2024sea} and may contribute to an increase in extreme weather events in mid-latitude~\cite{overland2021intermittency}. Additionally, sea ice loss affects the Arctic's role in regulating global ocean circulation, particularly the meridional overturning circulation (AMOC)~\cite{van2024role}, which has profound implications for global climate stability. Therefore, understanding and forecasting the seasonal variations of Arctic sea ice is not only important for regional climate assessments but also for grasping its global climatic implications. Accurate seasonal forecasts can help Arctic policymakers address climate change, promote sustainable resource development, and protect wildlife~\cite{lannuzel2020future}. They are also critical for understanding long-term climate dynamics, as the changes in the Arctic climate underscore the importance of precise monthly to seasonal projections for global climate systems~\cite{jung2016advancing}. However, the rapid changes in sea ice under global warming conditions pose challenges for forecasts, making the development of forecast methods both scientifically and practically significant.

Arctic sea ice forecasting research can be divided into physics-driven and data-driven approaches. Physics-based methods utilize numerical models integrating ocean, atmosphere, land, and ice interactions to understand the involved physical mechanisms~\cite{adcroft2019gfdl, horvat2017frequency, smith2016sea}. These models offer high precision but require significant computational resources. They are sensitive to initial conditions, and structural model physics errors can lead to systematic biases~\cite{blanchard2015model}. In contrast, data-driven approaches, including statistical and machine learning models, provide computationally lighter alternatives. Historically, data-driven sea ice forecasts have primarily targeted predictions of integrated Pan-Arctic September sea ice extent, achieving notable success in seasonal outlooks~\cite{schroder2014september, petty2017skillful, gregory2020regional}. While these extent-based forecasts demonstrate high skill metrics, their limited operational utility, stemming from coarse spatial resolution and inability to resolve regional ice dynamics, has motivated a paradigm shift toward spatially explicit sea ice concentration (SIC) and probabilistic forecasts. Statistical models such as vector autoregressive and Markov models are used in sea ice forecasting but struggle with capturing nonlinear dynamics~\cite{wang2016predicting, yuan2016arctic, wang2022reassessing, wang2019subseasonal, wang2023understanding, guemas2016review}. Machine learning, particularly deep learning (DL), has shown greater effectiveness in modeling the nonlinear evolution of sea ice. Current DL models forecast sea ice concentration (SIC) at both monthly~\cite{chi2017prediction, kim2018satellite, kim2020prediction, chi2021two, andersson2021seasonal, zhu2023deep} and daily~\cite{choi2019artificial, fritzner2020assessment, liu2021short, ren2022data, ren2023predicting, zheng2022mid, zheng2024spatio, grigoryev2022data} scales, offering forecasting capabilities for different time horizons. Despite these advancements in DL models, challenges remain in capturing multiscale dependencies in sea ice dynamics, which exhibit complex spatiotemporal patterns. Existing DL models, such as Convolutional Neural Networks (CNNs) and Long Short-Term Memory networks (LSTMs)~\cite{graves2012long}, struggle with spatial variability due to fixed receptive fields, which becomes particularly problematic in modeling complex, multiscale phenomena like sea ice dynamics. While Transformers~\cite{vaswani2017attention} capture a global receptive field, their quadratic complexity makes them inefficient for processing such large-scale, heterogeneous datasets. Moreover, the large volume of sea ice, ocean, and atmospheric data from diverse sources—such as observations, reanalysis, and modeling—introduces significant variability in data quality, which is strongly correlated with forecast accuracy. This heterogeneity in data sources often increases model uncertainty, complicating predictions.

To address these challenges, we introduce IceMamba, a novel DL architecture for seasonal SIC forecasting, combined with the State space model. To our knowledge, this is the first study to apply state space models to sea ice forecasting. IceMamba processes historical sea ice concentration (SIC) data along with reanalysis variables from ERA5~\cite{hersbach2020era5} and ORAS5~\cite{zuo2019ecmwf} to forecast monthly averaged SIC maps over the next several months at a 25 km resolution. We design the Residual Efficient State Space Block (RESSB) to enhance the model's performance by focusing on temporal and variable domains, enabling efficient extraction of correlations between climate and sea ice variables. In the spatial domain, RESSB inherits the global receptive field from the Vision state space block (VSSB)~\cite{gu2023mamba}, enabling IceMamba to efficiently capture the spatiotemporal dependencies of sea ice dynamics across different spatial regions, surpassing traditional DL models with fixed receptive fields. Moreover, IceMamba also inherits the linearly increasing computational complexity from VSSB~\cite{gu2023mamba}, making it more efficient than Transformers, which exhibit quadratic complexity as the input size grows. In the temporal and variable domains, RESSB effectively enhances IceMamba's attention to various input variables and their durations through the Efficient Channel Attention (ECA) module~\cite{wang2020eca}. This enables the IceMamba efficient extraction of correlations between different climate variables, thereby improving the forecast performance and stability of the model. Experimental results demonstrate that incorporating oceanic heat content and mixed-layer depth measurements from the ORAS5 reanalysis dataset significantly enhances IceMamba’s sea ice forecasting performance, establishing subsurface oceanographic parameters as critical predictors beyond surface-level sea surface temperature. Under the seasonal forecast benchmark \cite{bushuk2024predicting}, IceMamba outperforms all tested models in average RMSE and ACC metrics, while ranking second in IIEE. Through comprehensive evaluation of extreme September sea ice events, IceMamba demonstrates robust SIC forecast capabilities under critical conditions in the seasonal forecast benchmark comparisons.‌ Besides, the permutation-based explainability framework demonstrates IceMamba’s ability to dynamically prioritize key mechanisms across seasonal timescales: short-term forecasts depend predominantly on recent SIC, reflecting persistence, whereas longer lead times shift toward historical SIC from the same month in prior years, indicating an implicit seasonal alignment without explicit temporal encoding. The model also captures ice-albedo feedback through heightened sensitivity to upward solar radiation during summer melt, contrasting with weaker responses to downward radiation, while distinguishing atmospheric drivers from delayed oceanic influences. Detrending analyses suggest that apparent stratospheric linkages via 10 hPa zonal wind speed primarily arise from anthropogenic trends~\cite{kim2014weakening,xu2024influence} rather than direct dynamical coupling, as sensitivity to detrended 10 hPa zonal wind speed declines sharply. These results highlight IceMamba’s capacity to disentangle timescale-dependent interactions from rapid atmospheric forcing to slow oceanic adjustments, while maintaining physical consistency in feature attribution.

\section{Result}\label{sec2}

\bmhead{Overall Performance of IceMamba}
\begin{table}[h]
\caption{\label{tab:1} Comparison of mean MAE, RMSE, IIEE, and ACC for IceMamba variants and baseline models over non-land regions in the Pan-Arctic.}
\begin{tabular*}{\textwidth}{@{\extracolsep\fill}lccccc}
\toprule
 & MAE (\%) & RMSE (\%) & IIEE ($\times 10^{6}~km^{2}$)  & ACC \\
\midrule
    IceMamba-1 & 1.8065 & 6.9297 & 0.7865  &  0.9834 \\
    IceMamba-1-ERA5 &  1.8283&  7.0564& 0.7929& 0.9829\\
    IceMamba-4 & 2.3024 & 8.8384 & 1.0244 & 0.9731 \\
    IceMamba-4-ERA5 & 2.3258 & 8.9152 & 1.0388   &  0.9727\\
    IceMamba-6 &  2.4431 &  9.2994&  1.0899    & 0.9706 \\
    IceMamba-6-ERA5 & 2.5025 & 9.4414 & 1.1123   & 0.9691\\
    IceMamba-6-VSSB &  2.4674 & 9.435  & 1.0945 &  0.9696  \\
    Anomaly Persistence & 3.9940 & 11.1576 & 1.4653   &  0.9579\\  
\botrule
\end{tabular*}

\end{table}

\begin{figure*}
\includegraphics[scale=0.23]{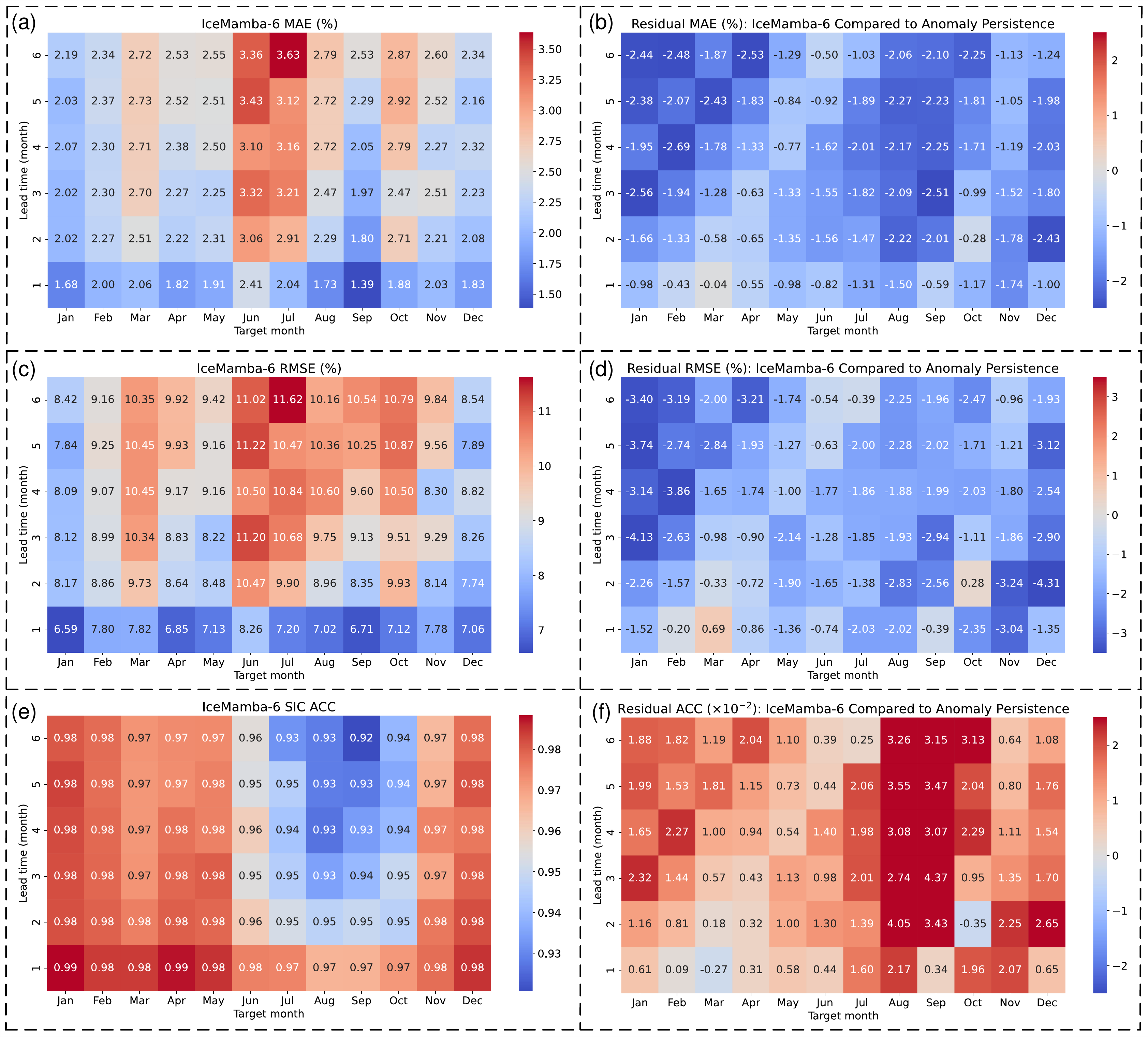}
\caption{\label{fig:2}Comparison of IceMamba-6 and Anomaly Persistence. (a), (c), and (e): MAE, RMSE, and ACC of IceMamba-6, averaged over the test years (2016-2022), presented for each target month (month of prediction) and lead time, with heatmap values in each grid cell. (b), (d), and (f): Heatmaps showing the differences in MAE, RMSE, and ACC between IceMamba-6 and Anomaly Persistence. All metrics are calculated over non-land regions in the Pan-Arctic.}
\end{figure*}

To evaluate the overall performance of the IceMamba architecture, we trained three variants (IceMamba-1, IceMamba-4, and IceMamba-6) using a consistent training scheme. Each variant is designed to forecast the monthly mean SIC for the next 1, 4, and 6 months, representing short-term (IceMamba-1), seasonal (IceMamba-4), and long-term seasonal forecasts (IceMamba-6), respectively. For each variant, we tested multiple input combinations (as shown in Supplementary Tables 2 to 4) and the data combination with the best forecast performance is selected as the final training set for IceMamba-1, IceMamba-4, and IceMamba-6. Additionally, we trained IceMamba-1-ERA5, IceMamba-4-ERA5, and IceMamba-6-ERA5 using only ERA5 data to compare the forecast performance of models only trained with SIC and ERA5 reanalysis variables. This comparative analysis delivers a comprehensive assessment of model performance with various forecast windows, while also delineating the specific contributions of ORAS5 reanalysis data in enhancing predictive capabilities at different temporal scales.

The optimal data configurations for these variants revealed a dependence on the scale of the forecast window and the oceanic processes involved. Specifically, IceMamba-1 achieved the best performance with a combination of ERA5, ohc300, and mld001 reanalysis data, reflecting the dominance of surface ocean dynamics in short-term predictions. Meanwhile IceMamba-4, optimized for seasonal predictions, performs better with a combination of ERA5, ohc300, and a deeper mixed layer (mld003), indicating the increasing importance of vertical oceanic processes over a longer forecast horizon. For IceMamba-6, the combination of ERA5, ohc700, and both shallow and deep mixed layer depths (mld001 and mld003) emerged as the optimal configuration, suggesting that deep ocean heat content and vertical mixing become critical for long-term seasonal predictions. These findings highlight the model’s ability to capture the shift from surface-driven to deeper ocean-driven processes as the forecast window lengthens. The scale-dependent shift in optimal input variables underscores the model’s sensitivity to oceanic climate mechanisms and provides insight into how different ocean layers influence sea ice variability at different temporal scales.

Table~\ref{tab:1} presents the Mean Absolute Error (MAE), Root Mean Squared Error (RMSE), Integrated Ice Edge Error (IIEE), and Anomaly Correlation Coefficient (ACC) for the test sets of each model, with all metrics calculated for the pan-Arctic non-land region. Comprehensive descriptions of these metrics are provided in the Supplementary Information. As shown in Table~\ref{tab:1}, IceMamba variants incorporating ocean reanalysis data exhibit consistent yet modest superiority across all metrics compared to their ERA5-only counterparts. Supplementary Figure 1 presents the error differences between the IceMamba variants and their corresponding ERA5 models. By comparing the various IceMamba variants with their corresponding ERA5 models, we notice that the benefit of including deep ocean parameters becomes more pronounced with longer forecast windows. While IceMamba-1 and IceMamba-4 also show improvements over their ERA5 counterparts in several months, the most consistent and significant improvement is observed for IceMamba-6 over the majority of the calendar year. This suggests that for longer-range forecasts, the additional deep ocean data provide enhanced initial conditions and better capture the evolving ocean-ice interactions. Moreover, IceMamba-6 demonstrates a notable reduction in forecast errors during the August-October period compared to other months. This suggests the presence of key ocean-ice coupling mechanisms during the critical transition from the melt season to freeze-up in the Arctic. During this seasonal window, enhanced ocean heat flux from subsurface warm layers and deepening of the mixed layer play a crucial role in influencing sea ice dynamics. These processes are more effectively captured by the oceanographic parameters in the ORAS5 reanalysis, which contribute to the improved model performance during this transitional phase. This result aligns with previous studies~\cite{bushuk2017skillful,blanchard2011influence,ordonez2018processes}, which demonstrated that the heat capacity of a deep mixed layer enables summer and spring SST anomalies to reemerge in late fall, thereby improving the predictability of SIE anomalies in this region. Given the strong linkage between SIE and SIC, our findings suggest a similar impact on SIC predictability.

To evaluate RESSB, we established IceMamba-6-VSSB as the baseline by substituting RESSB with two VSSBs in IceMamba-6. Both architectures maintained identical experimental configurations and datasets for comparative validity. Furthermore, the Anomaly Persistence forecast is chosen as a baseline comparison model. Anomaly Persistence is defined as the sum of the SIC anomaly and the climatology at each lead time. The climatology is computed using a 10‐year sliding window preceding the forecast to account for its temporal changes \cite{zampieri2018bright, liu2021extended, wei2022prediction}. Compared to the baseline models, all IceMamba variants outperform the Anomaly Persistence model, exhibiting lower MAE, RMSE, and higher ACC, indicating improved model fitting and reduced mean forecast errors. Furthermore, all IceMamba variants demonstrate lower IIEE, suggesting superior sea ice edge forecasting capabilities relative to the baseline models. Notably, IceMamba-6 outperforms IceMamba-VSSB across all metrics, highlighting that RESSB more effectively captures relevant variables and spatiotemporal features from the input data than VSSB.

Fig.~\ref{fig:2}(a), (c), and (e) display the average MAE, RMSE, and ACC for the IceMamba-6 model across each lead time and target month, respectively. As lead time increases, forecast errors for each target month gradually rise, while the ACC declines. Compared to other target months, the forecast errors and ACC from June to October exhibit more pronounced fluctuations. We hypothesize that this increase in errors is primarily due to the melting season of Arctic sea ice (June to September). During this period, rising temperatures lead to rapid ice melting, resulting in significant changes in ice coverage and dynamics. The pronounced variability in sea ice makes accurate forecasts more challenging. Moreover, the transition to autumn brings about complex interactions between melting ice, ocean currents, and atmospheric conditions, further exacerbating uncertainties in model forecasts. While incorporating ORAS5 subsurface ocean data reduces August-October forecast errors compared to ERA5-only models (Supplementary Fig.1), the error levels during this period remain higher than in other months.

As shown in Fig.~\ref{fig:2} (b) and Fig.~\ref{fig:2} (d), IceMamba-6 exhibits superior forecast performance, as evidenced by its consistently lower average MAE and average RMSE in almost all lead times and target months compared to the anomaly persistence model. Fig.~\ref{fig:2} (f) shows the ACC residuals between IceMamba-6 and the Anomaly Persistence model. The IceMamba model exhibits a more pronounced ACC decline with increasing lead time in August and September. However, IceMamba shows a significantly higher ACC during the melting season, indicating that IceMamba retains a significant predictive advantage in this period, which may reflect its ability to capture complex meteorological patterns that are not as effectively represented by simpler persistence models. 

\bmhead{Performance in Sea Ice Edge Forecast}

\begin{figure*}[ht!]
\centering
\includegraphics[scale=0.25]{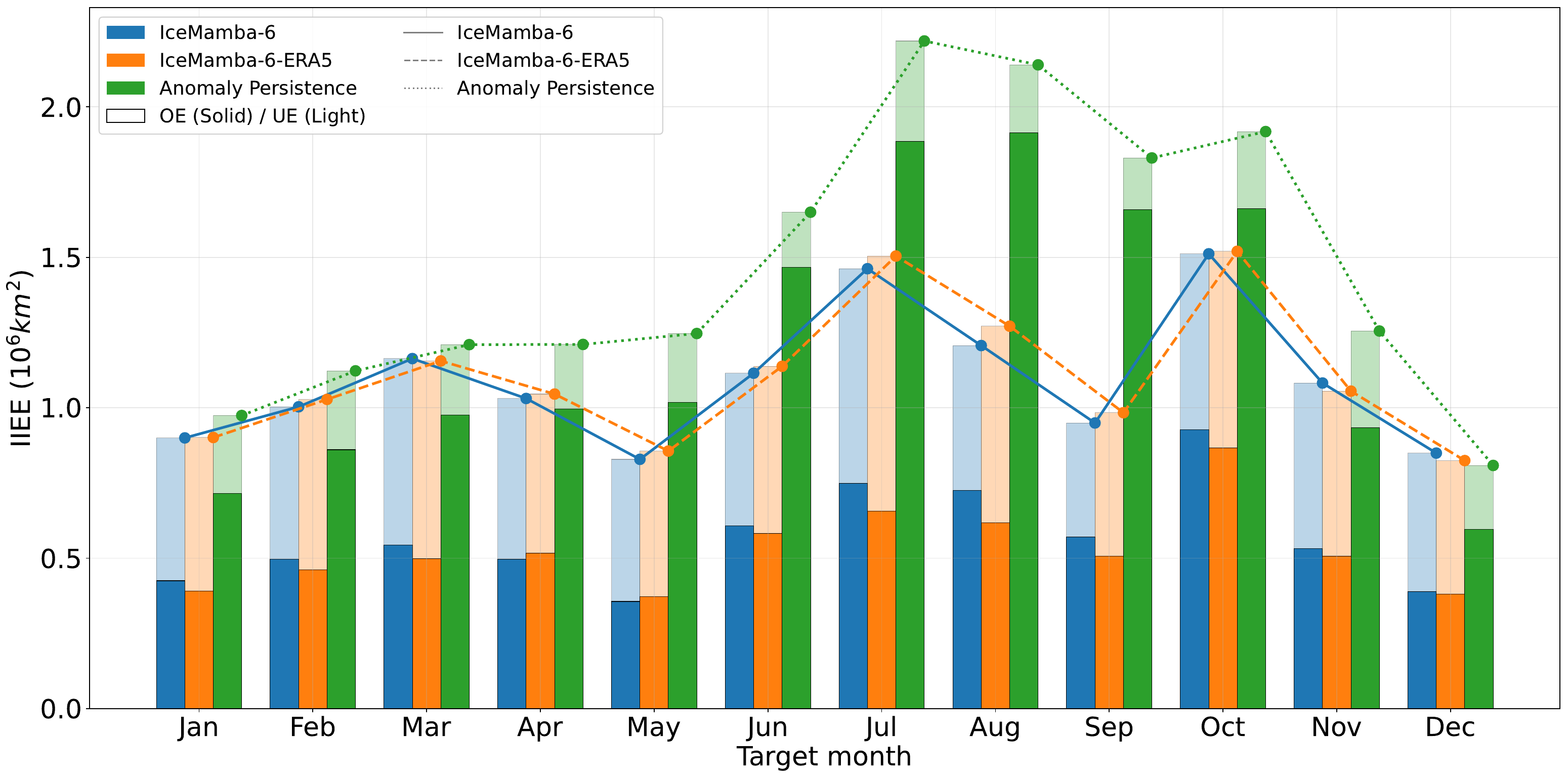}
\caption{Seasonal cycle of IIEE and its two components, overestimated error (OE) and underestimated error (UE), during the test period from sea ice forecast of IceMamba-6, IceMamba-6-ERA5, and Anomaly Persistence, averaged over six lead times.
}
\label{fig:3}
\end{figure*}

\begin{figure*}[ht!]
\centering
\includegraphics[scale=0.305]{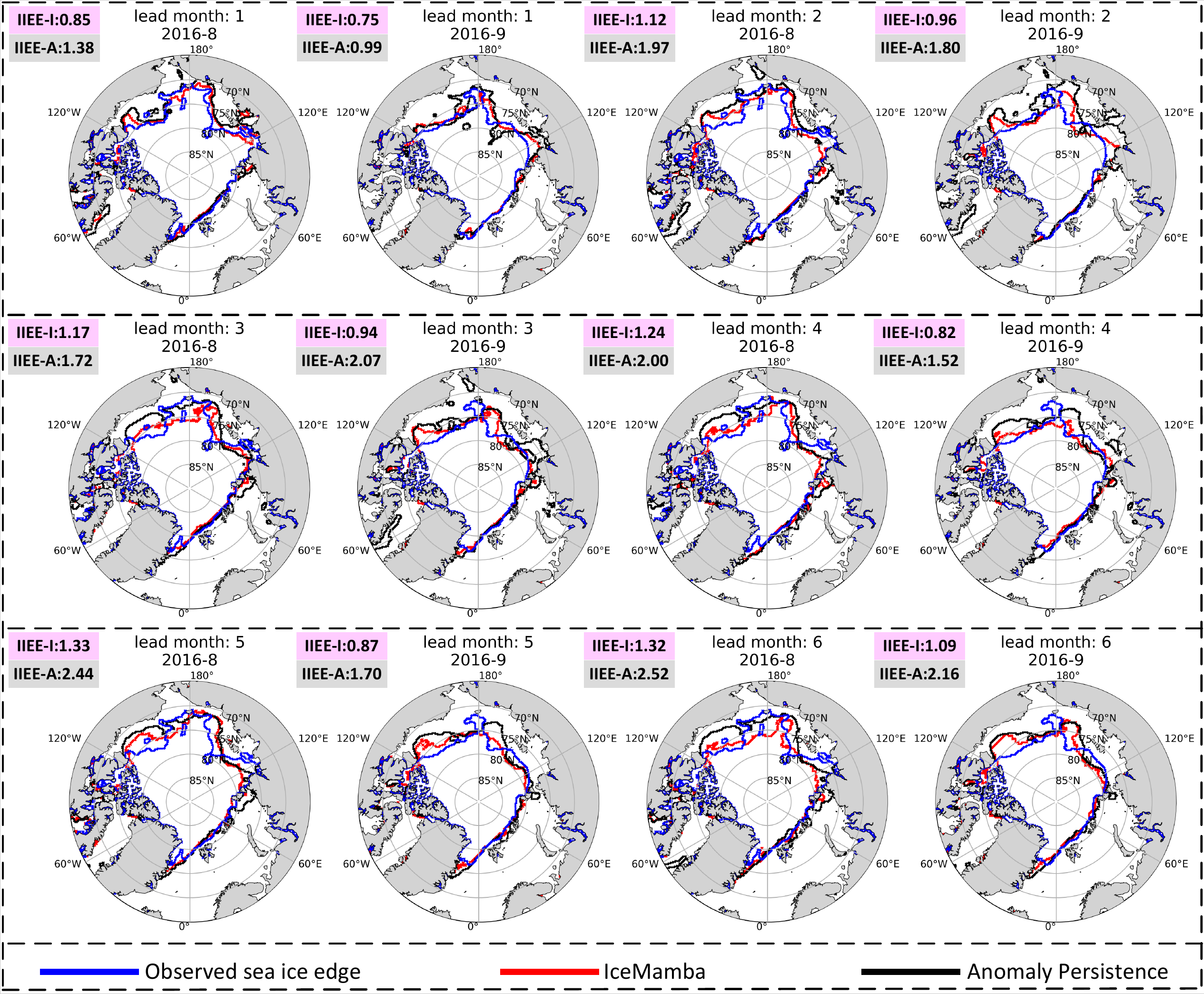}
\caption{Comparative visualization of predicted sea ice edge boundaries (colored contours) and Integrated Ice-Edge Error (IIEE) for August-September 2016 forecasts generated by IceMamba-6 (red) and the Anomaly Persistence benchmark (black), evaluated across 1 to 6 month lead times.}
\label{fig:4}
\end{figure*}

Figure~\ref{fig:3} illustrates the seasonal cycle of IIEE and its components (OE and UE) for IceMamba-6, IceMamba-6-ERA5, and Anomaly Persistence models, averaged across six lead times, with each value corresponding to the target month. Compared to the period from November to April, the IIEE of Anomaly Persistence has increased significantly from July to October. It may be caused by the complexity of the sea ice system from June to October. Sea ice is thinner between June and September and is more susceptible to atmospheric influences, significantly increasing uncertainty about sea ice changes.\cite{holland2006future,stroeve2012arctic}. From December to May, the sea ice is thicker and more stable, offering greater resistance to dynamical and thermodynamical effects, resulting in relatively minor interannual changes in the sea ice edge, making the IIEE of Anomaly Persistence relatively lower during this period. The IceMamba framework (IceMamba-6 and IceMamba-6-ERA5) shows a similar seasonal forecast pattern to Anomaly Persistence. However, IceMamba framework demonstrates significantly lower mean IIEE than Anomaly Persistence, particularly in summer and autumn. This improvement suggests that IceMamba framework effectively learns the complex, nonlinear relationships between SIC and reanalysis variables, providing a more accurate forecast of the sea ice edge during periods of higher variability. Moreover, the OE and UE distributions differ notably between the two models. Anomaly Persistence consistently shows significantly higher OE than UE throughout all 12 months, indicating a tendency to overestimate SIE (a limitation that could be partially mitigated by incorporating a damped anomaly persistence approach~\cite{bushuk2024predicting}).  In contrast, the IIEE of IceMamba framework aligns with seasonal variations, with OE and UE balancing more closely and adjusting to seasonal ice changes, providing a less biased but seasonally variable error distribution. 

Comparative analysis between IceMamba-6 and IceMamba-6-ERA5 shows that IceMamba-6, which integrates mixed layer depth (mld) and ocean heat content (ohc), shows a more noticeable reduction in the IIEE during July–October compared to other months. Although IceMamba-6 shows slightly lower IIEE values than IceMamba-6-ERA5 in a few months, its overall average IIEE remains lower. This performance gap highlights the crucial regulatory role of subsurface oceanic processes in Arctic sea ice forecast: Autumn mixed-layer deepening facilitates vertical heat transport from subsurface reservoirs, and accumulated ohc in marginal ice zones during summer creates delayed thermal forcing on basal ice melt. In contrast, the sst-restricted model fails to capture sub-mixed-layer heat flux processes, resulting in a systematic overestimation of sea ice extent during the late melt season. These findings quantitatively validate the necessity of incorporating subsurface ocean thermodynamics in sea ice forecasting systems.

Fig.~\ref{fig:4} compares the performance of IceMamba-6 and Anomaly Persistence in forecasting sea ice boundaries for August and September 2016. The results indicate that IceMamba significantly reduces the mean Integrated Ice Edge Error (IIEE) during peak melting periods, demonstrating enhanced accuracy in capturing ice edge position and morphological changes. The IIEE of both methods increases with lead time; however, the error growth rate of Anomaly Persistence is noticeably higher. The figure also reveals that for extended forecast periods, Anomaly Persistence often exhibits substantial deviations of the ice edge from observations, whereas IceMamba more effectively tracks the actual ice retreat, leading to lower overall errors.  In summary, the experimental results in Fig.~\ref{fig:4}  confirm IceMamba’s advantage in reducing mean IIEE and highlight its ability to maintain lower errors in long-term forecasts and capture local details demonstrating improvements in sea ice boundary prediction.

\bmhead{Skill comparison with dynamical and statistical models}
\begin{figure}
\centering
\includegraphics[scale=0.6]{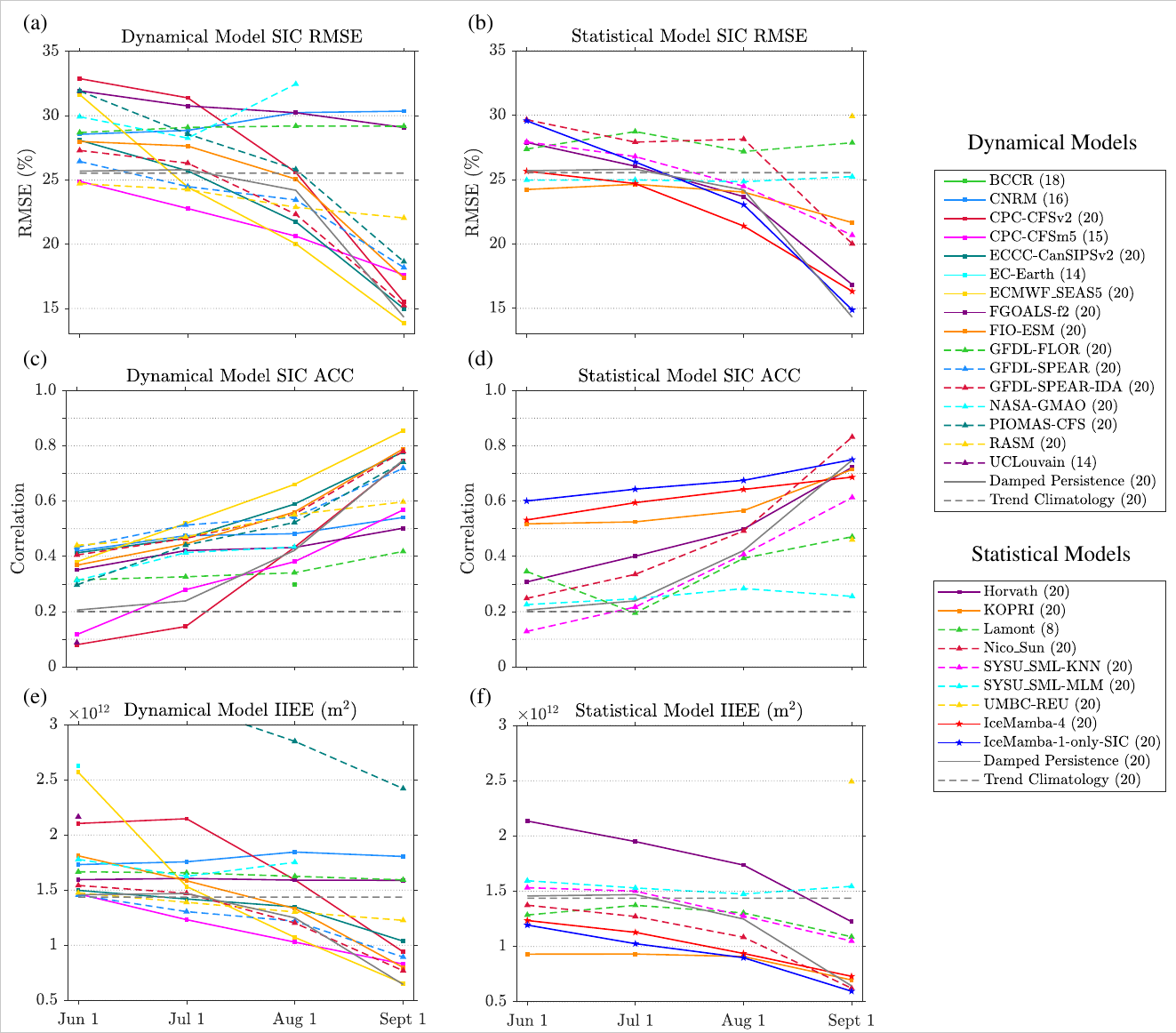}
\caption{RMSE, ACC, and IIEE for September SIC forecast (2001–2020). (a), (c), (e) show results for dynamical models, and (b), (d), (f) for statistical models. RMSE and ACC are averaged over regions where SIC standard deviation $> 10\%$. Models are color-coded, with reference forecasts in grey. Skill metrics are shown for each initialization from June 1 to September 1. Bracketed numbers in the legend indicate the years of data each model contributed over 20 years. \label{fig:5}}
\end{figure}

\begin{figure}
\centering
\includegraphics[scale=0.38]{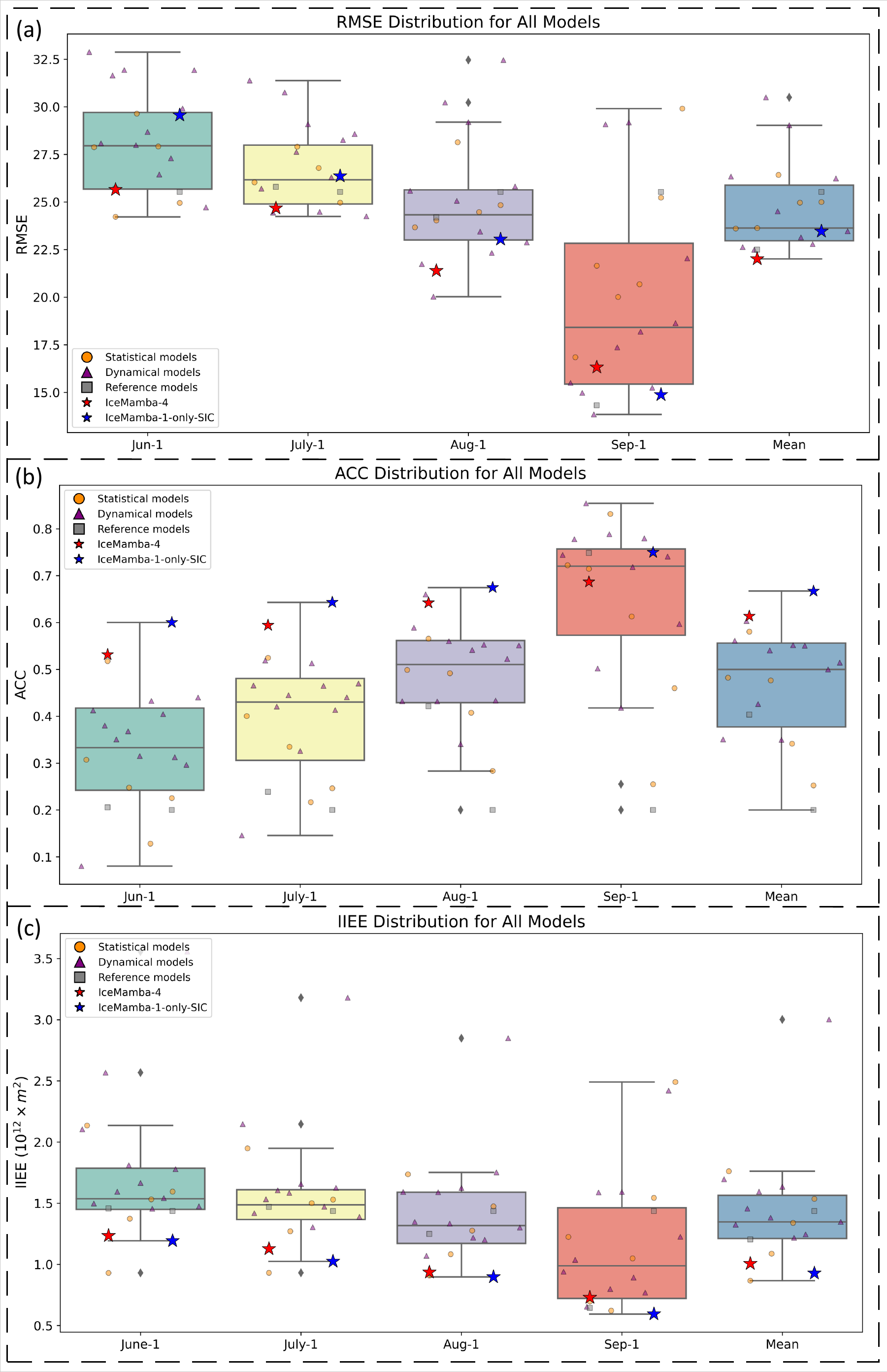}
\caption{Box plot illustration: RMSE, ACC, and IIEE for September SIC forecast (2001–2020) from models that contribute a full 20-year forecast. Panels (a), (b), and (c) show RMSE, ACC, and IIEE across models, averaged over regions with SIC standard deviation $> 10\%$. Models are color-coded: grey (reference), orange (statistical), and purple (dynamical). IceMamba-4 and IceMamba-1-only-SIC are highlighted in red and blue, respectively.
\label{fig:6}}
\end{figure}

To comprehensively assess the forecast performance of the IceMamba architecture, we adopted the evaluation benchmark proposed by \cite{bushuk2024predicting}, facilitating a systematic evaluation of various dynamical, statistical (including DL models) for September Arctic sea ice. Evaluation metrics for assessing SIC forecast performance include RMSE, ACC, and IIEE. RMSE and ACC metrics focus on regions where the standard deviation of September SIC exceeds 10\%, as this region exhibits significant variability, making it ideal for evaluating model performance.

To match the benchmark, IceMamba-4 and a retrained version of IceMamba-1, named IceMamba-1-only-SIC (using only SIC as input), were chosen for skill evaluation. IceMamba-1-only-SIC employs a recurrent-based forecast method. Specifically, with a forecast initiation date of June 1, 2001, the monthly mean SIC from June 2000 to May 2001 is used to predict the monthly mean SIC for June 2001. This prediction is then incorporated into the input data to forecast the monthly mean SIC for July 2001, and the process is repeated sequentially to forecast the monthly mean SIC for September 2001. while IceMamba-4 can directly predict the monthly mean SIC for each of the next four months in a single step. Moreover, both models undergo annual recalibration through a temporally segregated rolling-window framework. Let Y denote the target prediction year. The training set spans January 1979 to December of year Y-5 (e.g., 1979-1996 for Y= 2001), while the validation set occupies January Y-4 to December Y-1 (1997-2000 for Y= 2001). When forecasting for year Y + 1, the training window extends to December Y-4 (1979-1997 for Y+1= 2002), with the validation window advancing to span January January Y-3 to December Y (1998-2001). Crucially, the training window exhibits progressive expansion (annual increments), whereas the validation window maintains a fixed four-year span, ensuring both temporal isolation and evaluation consistency.

IceMamba-4 and IceMamba-1-only-SIC are benchmarked against 16 dynamical models \cite{kimmritz2019impact, voldoire2019evaluation, wang2013seasonal, saha2014ncep,  liu2019assessment, collow2020develop, lin2020canadian, hazeleger2012ec, johnson2019seas5, zuo2019ecmwf, li2021dynamical, qiao2013development, chen2016ocean, shu2021arctic, msadek2014importance, bushuk2017skillful, bushuk2022mechanisms, zhang2022subseasonal, molod2021gmao, zhang2003modeling, zhang2008ensemble, cassano2017development, barthelemy2018sensitivity}, 7 statistical models \cite{horvath2021linear, chi2017prediction, chi2021two, yuan2016arctic, bushuk2024predicting, lin2023optimization, zeng2023reducing, kim2021multi}, and 2 reference predictions (Damped Persistence \cite{van2007empirical}, Trend Climatology). Since the statistical models include several DL components, we also categorize IceMamba as part of the statistical models. The evaluation results for all tested models are illustrated in Fig.~\ref{fig:5}. Bracketed numbers in the legend indicate the years of data each model contributed over 20 years. All specific data of evaluation results for RMSE, ACC, and IIEE are recorded in Supplementary Tables 2, 3, and 4, respectively. 

As shown in Fig.~\ref{fig:5}, among the 25 evaluated models, 5 models did not provide complete forecasts from 2001 to 2020. To ensure a fair comparison, the RMSE, ACC, and IIEE distributions for the models that contributed complete 20-year forecasts are displayed in the box plots of Fig.~\ref{fig:6}. As shown in Fig.~\ref{fig:6}~(a), IceMamba-4 achieved the lowest average RMSE (22.0086\%) in all tested models. Moreover, only ECMWF\_SEAS5 (dynamic model, 22.4916\% RMSE) and IceMamba-4 exhibited lower errors than the damped persistence reference, highlighting their forecasting capabilities. For short-term forecasts (September 1 initiation), IceMamba-1-only-SIC attained the lowest RMSE among statistical models (14.8792\%), slightly surpassed by ECMWF\_SEAS5 (13.8444\%). Although dynamical models generally perform better in short-term forecasts compared to statistical models, IceMamba-1-only-SIC achieves competitive performance (15.2924\% RMSE), approaching ECMWF\_SEAS5 (13.8444\%). However, its reliance on a recurrent-based method leads to significant error accumulation as lead time extends, causing RMSE to rise sharply. Despite this limitation, its average RMSE (24.0335\%) remains competitive against non-IceMamba statistical models.

The Anomaly Correlation Coefficient (ACC) evaluation reveals that IceMamba-1-only-SIC achieves superior performance (mean ACC=0.6604), exceeding all comparative models (Fig.~\ref{fig:6}b). Notably, while IceMamba-4 incorporates additional climatic inputs, it demonstrates reduced ACC values (0.6126) relative to both IceMamba-1-only-SIC across all initialization dates, suggesting that multi-month SIC forecasting in a single step can compromise ACC. For IIEE, IceMamba-1-only-SIC achieves the lowest IIEE at September 1 ($0.6152 \times 10^{12}~m^{2}$) and August 1 ($0.9265 \times 10^{12}~m^{2}$) initializations, yet exhibits the second-lowest mean IIEE. Notably, IceMamba-4 maintains the third-lowest mean IIEE despite showing marginally higher values than IceMamba-1-only-SIC at all lead times. 

Overall, IceMamba achieves competitive performance compared to other tested models, particularly the IceMamba-4 variant, which achieves the lowest average RMSE in all evaluated models. For short-term forecasts initiated on September 1, IceMamba-1-only-SIC achieves the lowest RMSE among all statistical models, second only to the dynamical model ECMWF\_SEAS5. In terms of the ACC and IIEE evaluations, IceMamba-1-only-SIC performs well in the one-month lead forecats and IceMamba-4 maintains a clear advantage in overall performance. These findings collectively indicate that the IceMamba framework represents a significant advancement in the field of sea ice forecast, particularly in addressing the complexities and variability of Arctic climates.

\begin{figure*}
\centering
\includegraphics[scale=0.30]{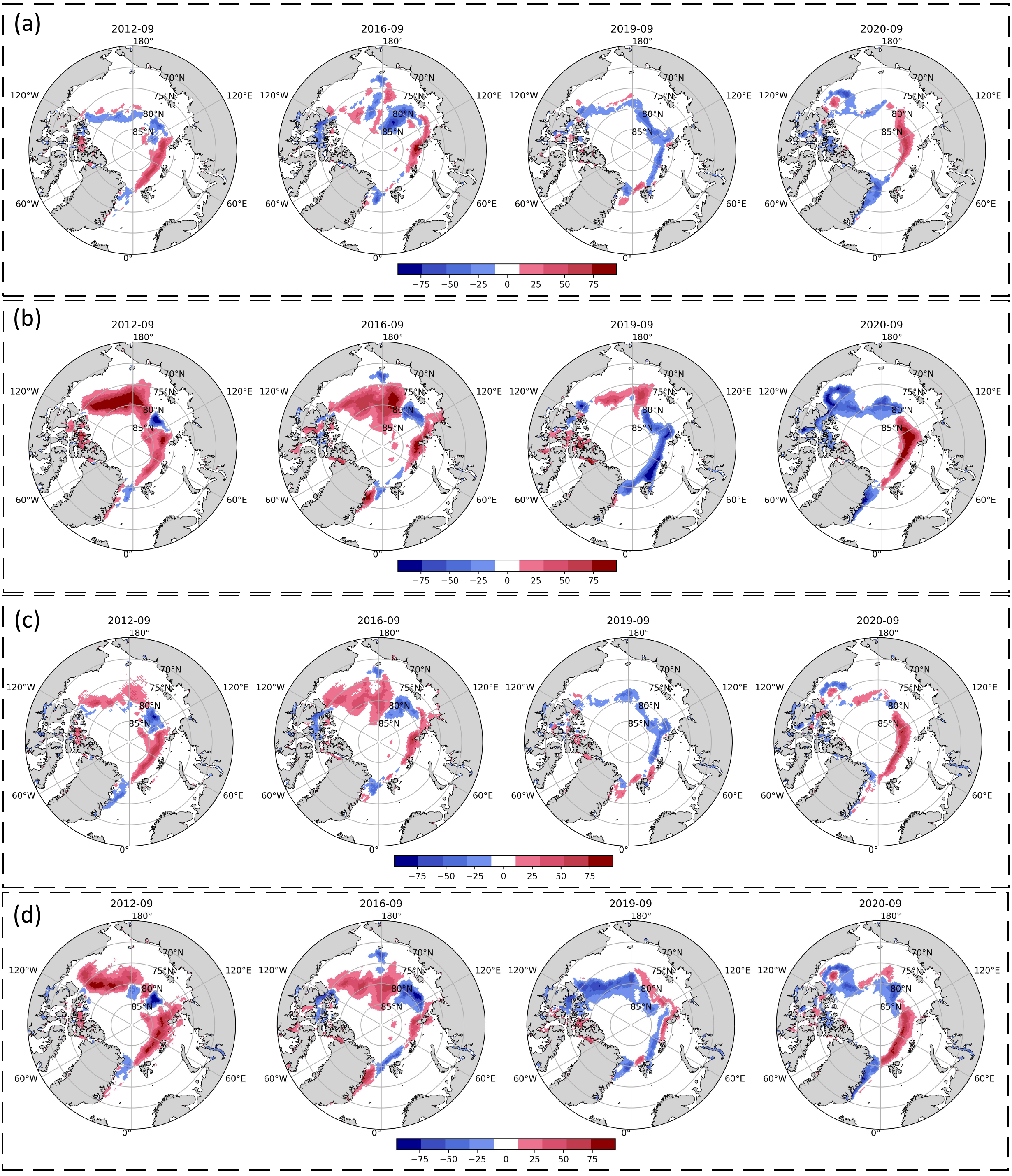}
\caption{ Residuals of SIC predictions.
(a-b) IceMamba-1-only-SIC at initialization dates September 1 and June 1;
(c-d) IceMamba-4 at initialization dates September 1 and June 1;
Residuals are calculated as model predictions minus NSIDC observational values. Color bars indicate residual magnitude, with blue/red tones representing systematic underprediction/overprediction respectively. 
\label{fig:7}
}
\end{figure*}

\bmhead{Forecast performance in extreme September sea ice events}

Arctic sea ice variability, especially during extreme low conditions at the end of summer, is a critical indicator of climate change. To better capture the early evolution of these anomalies, we include forecasts initialized on June 1, in addition to those on September 1, thus providing a more comprehensive evaluation of model performance during extreme sea ice events. Supplementary Tables 8 and 9 present a comprehensive comparison of IceMamba-4, IceMamba-1-only-SIC, statistical models, and dynamical models in predicting SIC during the extreme years (2012, 2016, 2019) based on the initialization date of September 1 and June 1, respectively. Since ACC requires data from multiple time points to effectively assess the correlation of anomaly patterns, calculating ACC at a single time point does not reflect the overall trend and consistency of model performance. Therefore we only compare the metrics of RMSE and IIEE in this Section. A further visualization of the performance of all evaluated models is provided by the box plot in Supplementary Fig.~2 and Supplementary Fig.~3, which illustrates the distribution of the RMSE and IIEE across the evaluated models. 

For the forecasts initialized on September 1, the results in Supplementary Fig.~2 (a) highlight the outstanding performance of the IceMamba models. IceMamba-1-only-SIC achieved the lowest RMSE in all non-IceMamba statistical models during each extreme year. Compared with dynamical models, only a few dynamical models reached lower RMSE than IceMamba-1-only-SIC. Notably, Dynamical models excel in short-term Arctic sea ice prediction primarily due to their capacity to assimilate near-real-time observational data into physically consistent frameworks. This synergy between data assimilation and physical principles ensures high-fidelity initial conditions, whereas purely data-driven approaches rely on historical training data that may not capture rapidly evolving initial states. In September 2019, IceMamba-1-only-SIC recorded an RMSE of 11.72\%, the best performance observed across all evaluated models. As shown in Supplementary Fig.~2 (a) (b), the IIEE metrics corroborate the effectiveness of the IceMamba. IceMamba-1-only-SIC achieved the lowest IIEE in most extreme years, significantly outperforming many other models, thereby reinforcing its reliability in predicting the sea ice edge. In comparison, IceMamba-4, which forecasts monthly averaged SIC maps over the next four months in one step, demonstrates higher IIEE than IceMamba-1-only-SIC. This difference can be attributed to the increased uncertainty inherent in long-term forecasts.

For forecasts initialized on June 1, a notable performance divergence emerged between the two model architectures. As shown in Supplementary Fig.~3 (a) and (b), IceMamba-1-only-SIC exhibited marked degradation in predictive skill. In contrast, IceMamba-4 maintained stable forecasting capabilities without significant performance decay. Particularly noteworthy are its achievements: attaining the lowest RMSE (18.73\%) in 2019 and simultaneously achieving both minimal RMSE and IIEE metrics in 2020 (20.92\%) among all dynamical and statistical models, demonstrating reliable forecast performance.
Notably, the Great Arctic Cyclone of August 2012 preconditioned extreme sea ice loss prior to the September initialization date. This implies that the initial conditions fed to the models already encapsulated extreme anomalies.
While IceMamba-1-only-SIC struggled to maintain accuracy under such preconditioned extremes - likely due to error accumulation in its recursive framework - IceMamba-4's superior performance suggests its architectural design effectively decouples initial condition dependencies through advanced feature extraction.

The residual of predicted and observed SIC is shown in Fig.~\ref{fig:7}. In general, the forecast errors in Sep 2012, 2019, and 2020 are mainly concentrated on marginal regions. Since 2012 and 2020 are the years with the lowest and second lowest SIE, respectively, our model generally overestimated SIC in these years. Comparative residual analysis between the two models shows nuanced temporal dependencies. When initialized on September 1, IceMamba-1-only-SIC demonstrates superior short-term predictive skill, evidenced by its narrower residual distribution range compared to IceMamba-4. This advantage, however, diminishes with extended lead times. Under June 1 initialization conditions, IceMamba-1-only-SIC exhibits significantly amplified residuals (indicated by intensified color gradients), while IceMamba-4 maintains relatively stable error patterns. This divergence is particularly pronounced in the extreme ice loss years of 2012 and 2020. Overall, these results indicate that the IceMamba models are proficient in forecasting SIC under extreme climate conditions. This capability is crucial for enhancing our understanding of Arctic dynamics. By accurately forecasting SIC, the IceMamba models provide a vital tool for researchers and policymakers aiming to formulate effective strategies in response to the evolving climate landscape.

\bmhead{Explainability of IceMamba}

\begin{figure}
\centering
\includegraphics[scale=0.3]{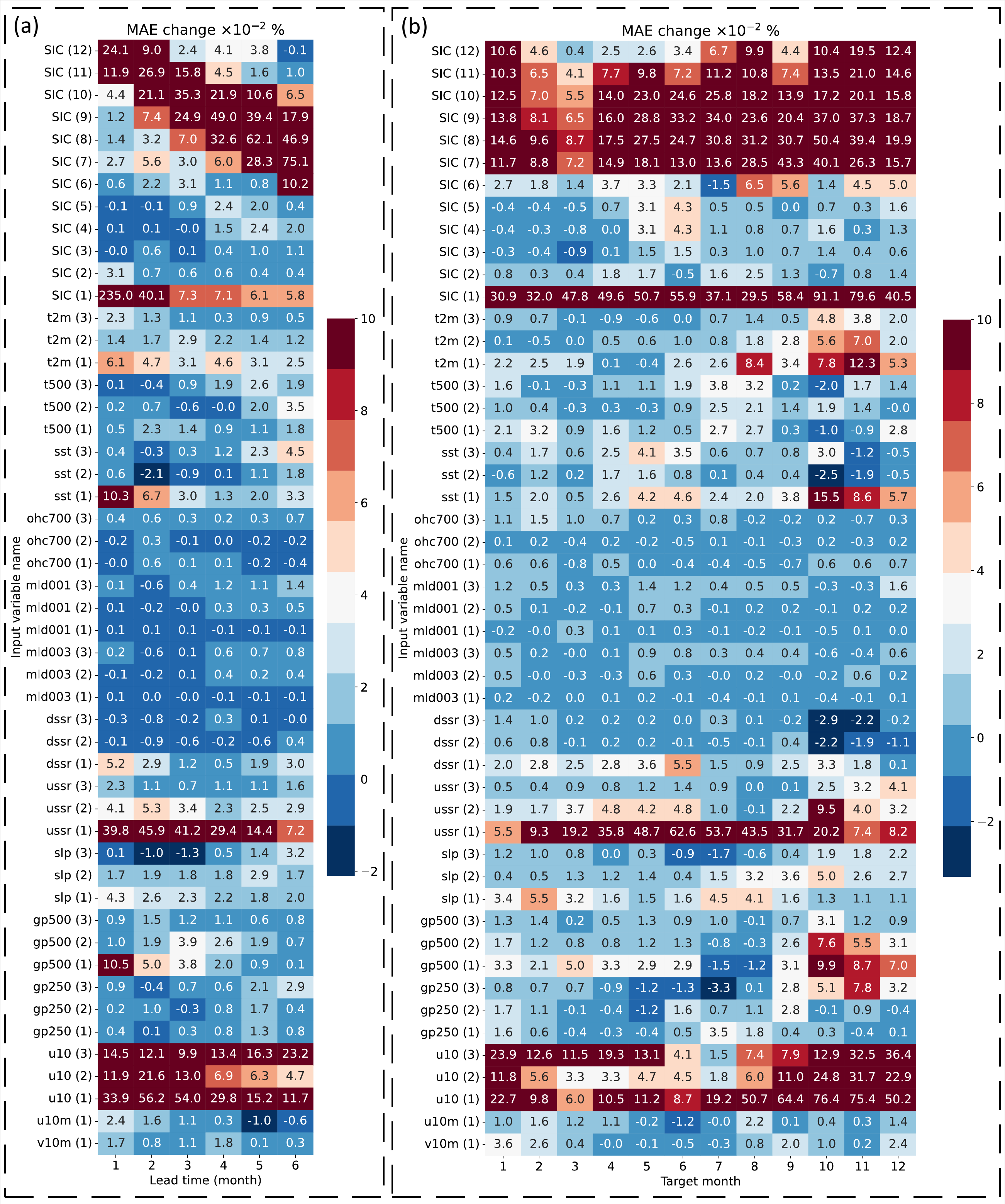}
\caption{\label{fig:8} Heatmaps from the permute-and-predict method: (a) Mean MAE change for each input variable in every lead time. (b) Mean MAE change for each input variable in every target month. MAE averaged over 10 random seeds and 84 forecast months (2016-2022). Bracketed numbers indicate the input lag (month)}
\end{figure}

Achieving Explainability in DL models is crucial for understanding the mechanisms behind their forecasts. We implement a permutation-based Explainability framework. For each variable, we randomly shuffle its 2D input fields across temporal dimensions while keeping other variables unchanged. The degradation in IceMamba's forecast skill (quantified by MAE increase) after permutation systematically reveals each variable's relative importance across forecast months and lead times. Fig.~\ref{fig:8} (a) and (b) present heat maps showing the changes in MAE after the permutation of specific variables. The y-axis denotes the types of permuted variables, with numbers in brackets indicating the input lag.

Among all variables, SIC (1), reflecting the SIC with one lag month, exerts the most substantial influence in one-month forecasts, highlighting the predictive significance of sea ice persistence for short-term changes. As lead time increases, reliance on recent SIC data diminishes, with the model increasingly utilizing last year’s SIC value for the same target month corresponding to the forecast month, demonstrating a clear seasonal alignment. The alignment is evident in the influence shift with lead time. At one-month and two-month lead times, SIC (1) is the most influential variable, while SIC (12) and SIC (11) are the second most influential variables, corresponding to the forecast target month for each respective lead time from the previous year. By three months, SIC (10) becomes the primary influence, while the effect of SIC (1) decreases. This pattern continues with longer lead times, where each step reflects a significant impact from the same month’s SIC in the previous year. This pattern demonstrates the model's capacity to implicitly capture and utilize the seasonal cycle of SIC by focusing on historical data relevant to the forecasted month. Notably, this seasonal alignment occurs without explicit temporal coding, suggesting the model has developed an internalized understanding of annual SIC variations. 

For most non-SIC climate parameters, the model exhibits maximum sensitivity to data from one lag month. However, u10 (zonal wind at 10 hPa) stands as an exception. Unlike near-surface wind parameters (u10m and v10m), which directly influence sea ice through surface drag, u10 cannot physically interact with sea ice. Despite this, u10 shows a strong and sustained influence across three lag months, particularly from July to December, and impacts all lead times. This unique sensitivity to u10 suggests the model’s potential recognition of stratospheric drivers. However, this apparent stratospheric linkage raises the question of whether it is primarily driven by physical mechanisms or by the climate change signal, with anthropogenic sea ice loss since 1979 weakening the stratospheric polar vortex~\cite{kim2014weakening, xu2024influence}.

To answer this question, We retrain the model using the original training data with only u10 detrended to remove long-term climate trends. The results in  Supplementary Fig.~4  showed that IceMamba's sensitivity to u10 significantly decreased after removing the trend, indicating that the relationship could be driven by shared anthropogenic trends, rather than direct causal effects of stratospheric wind. Notably, training on detrended anomaly ua10 led to a decline in model performance ( Supplementary Table. 10), indicating that long-term climate trends in ua10 provide valuable predictive signals. Overall, The sharp reduction in sensitivity to detrended u10 and the decline in predictive skill when using anomalies underscores the role of long-term human-driven sea ice loss and its cascading impacts on the stratospheric polar vortex. While the model successfully identifies shared trends between u10 and sea ice variability, this relationship mainly reflects a co-evolution of anthropogenic forcing rather than a mechanistic stratosphere-ice coupling. This highlights the critical importance of disentangling climate change signals from intrinsic dynamical processes in attribution studies. Furthermore, the retained predictive value of long-term trends suggests that anthropogenic forcing provides a key "memory" for seasonal sea ice forecasts, even as it complicates the interpretation of causal drivers in coupled climate systems.

ussr (1), representing upward surface solar radiation with one lag month, demonstrates a noticeable and consistent impact across all lead times, with a peak influence specifically during the Arctic summer months. This period corresponds to the seasonal peak in solar radiation and enhanced ice melt, suggesting that the model captures the role of radiative fluxes in driving the summer sea ice melt through the albedo effect. The model’s sensitivity to ussr (1) also reflects an understanding of the seasonal dependency of radiative forcing and its critical role in Arctic ice melt dynamics. In contrast, dssr representing downward surface solar radiation, has a weaker influence on sea ice prediction compared to ussr. This difference likely arises from how these parameters interact with ice-albedo feedback. While dssr provides the initial solar energy input, ussr captures the net radiative balance after reflection by ice. In the Arctic, high-albedo ice and snow reflect most of the incoming solar radiation. The ussr accurately measures this reflection process. Thus, the model’s heightened sensitivity to ussr suggests it recognizes the critical role of the albedo-driven feedback loop in amplifying seasonal ice melt.

For the atmospheric temperature variable t2m (1), representing 2m air temperature with a one-month lag period. its predominant influence on August-November forecasts demonstrates critical regulatory effects on autumn predictability. This aligns with the Arctic’s seasonal transition from ice-melt to freeze-up, during which delayed surface heat fluxes—driven by declining albedo and amplified ice-ocean-atmosphere feedbacks—propagate thermal anomalies into the boundary layer. In contrast, the model exhibits lower sensitivity to t500 relative to t2m, reflecting the model’s limited sensitivity to upper atmosphere temperature.

For ocean variables, sst (1), representing sea surface temperature with one lag month, has the strongest influence, particularly from October to December, suggesting that the model captures the thermal inertia of the ocean in delaying ice formation beyond the melt season. While the model has lower sensitivity to ohc and mld compared to sst and other atmospheric variables. Interestingly, the sensitivity to ohc and mld is mainly concentrated in the third lag month, which is significantly different from other variables. This delayed response reflects the deeper ocean memory, where subsurface processes, such as the release of accumulated heat from the deeper ocean layers and mixed-layer dynamics, gradually influence ice-ocean-atmosphere interactions. The model successfully distinguishes between the dynamics of subsurface processes and those of other variables, capturing the delayed effects of deeper oceanic memory. However, its relatively low sensitivity to these processes suggests that while the model acknowledges their importance, it faces challenges in fully capturing the slow, multi-month thermodynamic adjustments involved.

For pressure variables, gp500 (1), representing 500 hPa geopotential height with one lag month, demonstrates a notable effect across the one and two-month lead time during autumn and early winter (October to December). This pattern suggests that the model may implicitly recognize the role of mid-tropospheric pressure systems in modulating atmospheric circulation patterns, which can impact ice formation and stability during these months. In contrast, slp (1) and gp250 (1), representing sea level pressure and 250 hPa geopotential height with one lag month, respectively, show relatively weaker impacts, reflecting the model’s limited sensitivity to surface and upper-level pressure variations in shaping seasonal sea ice changes.

\section{Discussion}\label{sec3}

In this study, we introduce IceMamba, a novel DL architecture designed for the seasonal forecast of SIC over the pan-Arctic region. Experimental results indicate that the state space model is particularly well-suited for the field of sea ice forecast. The inherent capacity of Vision state space block (VSSB) to capture long-range dependencies enables them to effectively represent the complex dynamics of sea ice, which are influenced by an array of climatic factors over extended periods. This characteristic is essential for understanding the fluctuations and variability of sea ice in the Arctic, allowing for a more comprehensive analysis of the mechanisms driving these changes. The integration of the Residual Efficient State Space Block (RESSB) further enhances the forecast performance. By incorporating Efficient Channel Attention (ECA) and a residual branching structure, RESSB optimizes feature selection, enabling the model to prioritize the most relevant climatic signals. This optimization aligns well with the unique challenges of sea ice forecasting. To our knowledge, this is the first study to apply state space models to sea ice forecasting.

In a comprehensive comparison with dynamic and statistical models, IceMamba-4 achieved an average RMSE of 22.0086\%, positioning it as the leading model among all evaluated models.  For the short-term forecast, IceMamba-1-only-SIC recorded the lowest RMSE of 15.2924\% under the initialization date of September 1. ACC analysis revealed that IceMamba-1-only-SIC outperformed all models with a mean ACC of 0.6604, whereas IceMamba-4 had a slightly lower ACC of 0.6126, indicating the inherent trade-offs in forecasting multiple months consecutively. For IIEE, IceMamba-4 and IceMamba-1-only-SIC have lower average IIEE than all dynamical models. IceMamba-1-only-SIC had the lowest values at initialization dates of September and August, but its performance deteriorated significantly with longer lead times, reflecting a mean IIEE of $0.9459 \times 10^{12} \, m^{2}$, which was the second lowest among all evaluated models. 

IceMamba demonstrates robust forecasting capabilities for extreme Arctic September sea ice, with IceMamba-4 achieving the lowest RMSE (11.72\% in 2019) and stable performance under June-initialized long-term predictions, while IceMamba-1 excels in short-term accuracy. The ability to accurately forecast SIC, even during significant sea ice decline, makes IceMamba a vital tool for understanding Arctic dynamics and aiding researchers and policymakers in responding to climate change. Moreover, The interpretability experiments demonstrated that IceMamba retains a clear understanding of the relationships between input climate features and output predictions, confirming the model's efficacy in revealing the underlying mechanics driving sea ice dynamics. This transparency enhances the model's utility in both research and policy-making contexts.

Despite its strengths, this study still has some limitations. One limitation is that IceMamba adopts a channel fusion approach, which couples all input data into one dimension. Although RESSB can effectively capture the correlation between data from different channels, channel fusion requires all data to use the same grid method. This re-gridding process may introduce inherent biases in the data preparation process. Channel fusion also fixes the spatial range of climate data in the Arctic region, but the impact of atmospheric variables is global, which may also limit the prediction effect of IceMamba. Additionally, the spatial resolution of the input data (25 km) might not capture fine-scale processes that could influence sea ice dynamics, such as localized ocean currents and small-scale atmospheric phenomena. Another limitation pertains to the generalizability of the model. While IceMamba has shown commendable performance in the pan-Arctic region, its applicability to other regions, such as the Antarctic, remains to be tested.

To address these limitations, future research could explore alternative methods for data fusion that mitigate the biases introduced by re-gridding. Additionally, future work could investigate the incorporation of global atmospheric variables more explicitly into the IceMamba model. By integrating data that captures global atmospheric patterns and their influences on the Arctic, the model could potentially improve its forecast performance. Expanding the model to include physical processes explicitly, through hybrid modeling approaches that combine physical and data-driven methods~\cite{finn2023deep, finn2024generative, gregory2023deep, gregory2024machine}, thereby enhancing its robustness and interpretability. Testing IceMamba in different geographical contexts, such as the Antarctic, would provide insights into its versatility and potential adaptations required for different sea ice regimes.

\section{Method}\label{sec4}
\bmhead{Experimental data}

The datasets used in this research comprise observational SIC and observational climate reanalysis data. The SIC data is obtained from NOAA/NSIDC Climate Data Record of Passive Microwave Sea Ice Concentration version 4 product, which focuses on pan-Arctic region covering $448\times304$ grids (about $39.36^\circ$N–$89.84^\circ$N, $180^\circ$W–$180^\circ$E) with a spatial resolution of 25 km. This type of grid format is also known as the Equal Area Extensible Earth 2 (EASE2) Grid. The product provides passive-microwave-derived SIC estimates conforming to NOAA Climate Data Record (CDR) standards, combining the NASA Team (NT) \cite{cavalieri1984determination} and NASA Bootstrap \cite{comiso1986characteristics} (BT) algorithms, which estimate SIC from passive microwave brightness temperatures at various frequencies and polarizations. The CDR product refines these estimates by adjusting algorithm coefficients for each sensor, ensuring consistency in daily and monthly SIC time series. It began with NASA Nimbus-7 SMMR data in 1978, and brightness temperature input data are processed and incorporated into the record every three to six months. The reanalysis variables utilized in this study are derived from the European Centre for Medium-Range Weather Forecasts (ECMWF) ERA5 reanalysis at a resolution of $0.25^\circ$ \cite{hersbach2020era5}.  ERA5 covers the global climate from January 1940 up to the present day, offering detailed insights into atmospheric, land, and oceanic reanalysis variables on an hourly basis. We utilized ERA5 monthly averaged datasets spanning from 1979 to 2022 for surface variables on single levels and for upper air variables on pressure levels \cite{hersbach2019era5single, hersbach2019era5pressure}. Moreover, we incorporated ocean reanalysis data from the ORAS5 product to further investigate subsurface oceanic processes. ORAS5 provides a global ocean reanalysis, delivering monthly averaged fields over the period from 1958 to the present. The inclusion of ORAS5 data enables a more comprehensive analysis of the ocean’s thermal structure and its interactions with the atmosphere~\cite{zuo2019ecmwf}.

To focus on sea ice grid points, before fitting the SIC data into IceMamba, the SIC values in points of land are set to 0. In part of the SIC data, information around the North Pole is absent because of the "polar hole" phenomenon. The "polar hole" refers to a certain region around the geographic North Pole where satellite data coverage is limited by satellite orbits. As technology advances, the coverage of polar holes continues to decrease. The SMMR instrument exhibited a pole hole from November 1978 to June 1987, characterized by an area of $1.19 \times 10^6 km^2$ and a radius measuring 611 km, situated at a high latitude of $84.5^\circ$N. Subsequently, the SSM/I instrument featured a more modestly-sized pole hole, active from July 1987 to December 2007. This hole, with a radius of 311 km and a latitude of $87.2^\circ$N, covered an area of $0.31 \times 10^6 km^2$. Presently, the SSMIS instrument boasts the smallest pole hole to date, with a radius of 94 km, a latitude of $89.18^\circ$N, and an area of $0.029 \times 10^6 km^2$, in operation since January 2008 \cite{digirolamo2022sea}. The CDR addresses the polar "pole hole" issue by employing spatial interpolation. Initially, a temporal interpolation is applied using data from up to five days before or after to fill the pole hole locations. If such data points are unavailable, the pole hole is filled with the average SIC from surrounding grid cells.

All reanalysis datasets are re-gridded from a latitude-longitude grid to the EASE2 grid using bilinear interpolation. Due to the strong periodicity of meteorological data, some reanalysis variables were calculated as abnormal values to highlight the differences from climate values. These anomaly variables were determined during the training phase by deducting the climatological average for each respective target month from the observed values. In addition, we normalize the reanalysis variables by subtracting the mean and using the standard deviation calculated during the training year (1979-2010) to keep the values of each variable within a similar range, thereby improving the training stability of the model. The details of all input variables are shown in Supplementary Table 1.

\bmhead{Input variables of IceMamba}
Inspired by the parameter selection and climate preprocessing in \cite{andersson2021seasonal}, we adopted a subset of their methods, but IceMamba's input data excludes linear trends, land mask, and initialization date encoding (cosine and sine of the initialization date index). To address potential subsurface ocean influences, we further incorporated mixed-layer depth (mld) and ocean heat content (ohc) anomalies from reanalysis datasets, which effectively reflect heat fluxes below the mixed layer, critical for autumn sea ice variability. IceMamba uses 12 historical monthly mean SIC from NSIDC, and historical mean climate reanalysis sequences from ERA5~\cite{hersbach2020era5} and ORAS5~\cite{zuo2019ecmwf} for 1-month or 3-month lead times as input, capturing the spatiotemporal and ice-atmosphere-ocean coupling relationships within the sequence and generating predictions for the SIC over the next several months.

The input data includes six key groups: sea ice, atmospheric temperature, ocean, radiation, pressure, and wind. SIC represents the sea ice group. The atmospheric temperature group contains 2-metre air temperature (t2m) and 500 hPa air temperature (t500). The ocean group contains sea surface temperature (sst), Ocean heat content for the upper 300m (ohc300), Ocean heat content for the upper 700m (ohc700), Mixed layer depth 0.01 (mld001), and Mixed layer depth 0.03 (mld003). The radiation group contains upwards surface solar radiation (ussr) and downwards surface solar radiation (dssr). The pressure group contains 500 hPa geopotential height (gp500) and 250 hPa geopotential height (gp250). Finally, the wind group contains 10-metre X-direction wind speed (u10m), 10-metre Y-direction wind speed (v10m), and 10 hPa zonal wind speed (u10). Details of all input data are shown in Supplementary Table~1.

These climate variables are chosen to capture key processes affecting sea ice dynamics, which encompass both thermodynamic and dynamic processes. Thermodynamic processes involve the exchange of heat between the sea ice, atmosphere, and ocean, driving the growth and melting of sea ice. For example, the atmospheric temperature, radiation, and ocean groups capture the energy balance critical for seasonal melt and growth cycles. Dynamic processes involve mechanical forces that cause sea ice to move, deform, and redistribute. For instance, u10m and v10m in the wind group directly influence sea ice drift, while the 10 hPa zonal wind is included to account for teleconnections between the stratospheric polar vortex and Arctic sea ice anomalies. The pressure group captures large-scale atmospheric circulation patterns that drive ice motion and deformation. These input variables allow IceMamba to model the key mechanisms that influence sea ice.

To avoid the effect of higher uncertainties in the late 1978 observation, only SIC data from 1979 to 2022 are used in this study. The temporal coverage for the training, valid, and test sets, respectively, spans from Jan 1979 to Dec 2010 (32 years), Jan 2011 to Dec 2014 (4 years), and Jan 2015 to Dec 2022 (8 years). Note that IceMamba uses the monthly average SIC of the previous 12 months as input, the test period corresponding to the test set is 2016-2022.

\bmhead{IceMamba: A state space model for sea ice forecasting}

\begin{figure*}
\centering
\includegraphics[scale=0.131]{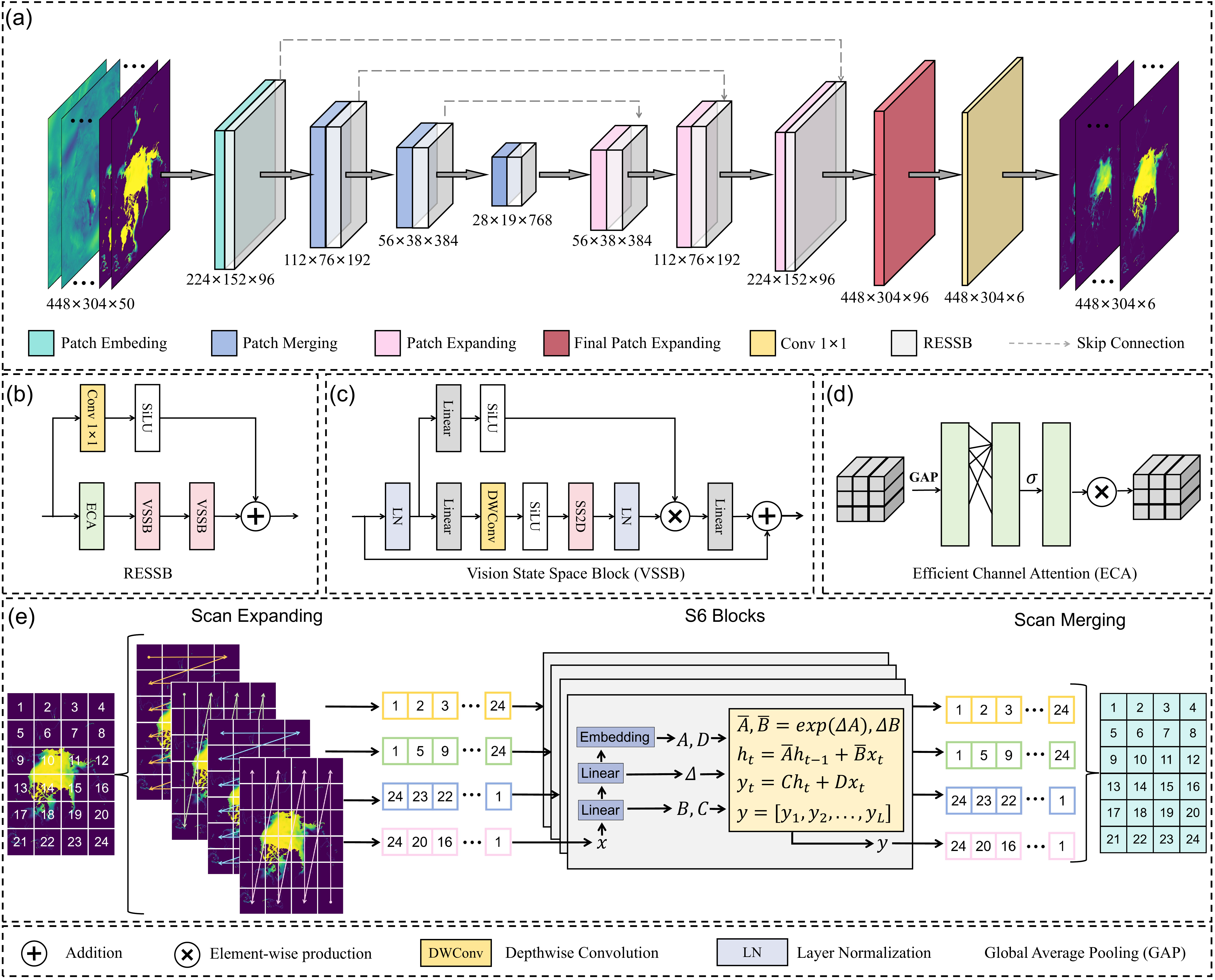}
\caption{\label{fig:1}Schematic diagram of IceMamba framework:
(a) Structure of IceMamba-6.
(b) Structure of Residual Efficient State Space Block (RESSB).
(c) Structure of Visual State Space Blocks (VSSB) in RESSB.
(d) Structure of Efficient Channel Attention (ECA) module in RESSB.
(f) Illustration of the 2D-Selective-Scan (SS2D) used in VSSB.
}
\end{figure*}

Recently, state space models (SSMs) have become a powerful tool for modeling continuous long sequences of data by representing system dynamics through latent variables. Mamba \cite{gu2023mamba} significantly advances SSMs with a selection mechanism that enhances efficiency and performance, enabling them to handle global contextual information effectively. Mamba models long-range dependencies by formalizing discrete state-space equations into a recursive form and combining them with a specially designed structured reparameterization, along with hardware-aware optimizations, making it a promising alternative to the Transformer \cite{vaswani2017attention}. Building on Mamba, Vmamba \cite{liu2024vmamba} is tailored for computer vision tasks. It addresses the direction-sensitive problem\cite{liu2024vmamba} by using the Cross-Scan Module (CSM) to capture interrelations among image patches, significantly enhancing performance in object detection \cite{chen2024changemamba,chen2024mim}, segmentation \cite{ma2024rs, liao2024lightm}, and classification \cite{chen2024rsmamba,yue2024medmamba}. VMamba excels in the efficient management of long-sequence data and the extraction of spatial and temporal features. Unlike CNNs, which primarily focus on local spatial information and struggle to capture long-range dependencies, Mamba seamlessly integrates these long-range dependencies with its advanced sequence processing capabilities. Furthermore, Mamba surpasses the traditional Transformer \cite{vaswani2017attention} model by optimizing computational efficiency, reducing memory usage, and maintaining high performance even with large datasets. Given that satellite and climate data are often presented in gridded formats akin to images, this makes VMamba particularly well-suited to address the complexities and demands of climate prediction tasks. 

Leveraging the potential of SSMs in satellite data processing, we propose IceMamba, a novel DL framework employing an Advanced SSM to forecast monthly mean SIC for the next several months. As illustrated in Fig.~\ref{fig:1}(a), IceMamba adopts an encoder-decoder framework with multiple Residual Enhanced State Space Blocks (RESSB). Each RESSB (Fig.~\ref{fig:1}(b)) integrates two Vision State Space Blocks (VSSB), a 1×1 convolutional layer, and an Efficient Channel Attention (ECA) module (Fig.~\ref{fig:1}(d)). ECA is added to RESSB to enhance the learning of different channel representations and select key channels to prevent redundancy. This enables IceMamba to focus on the key climate variables in the input data and prioritize the most influential lagged month of the corresponding climate variable. In addition, a residual branch with 1×1 convolution further improves pattern capture and information fusion. The VSSB (Fig.~\ref{fig:1}(c)) combines Layer Normalization \cite{ba2016layer}, Linear layers, Depthwise convolution \cite{chollet2017xception}, and a 2D-Selective-Scan (SS2D) block \cite{liu2024vmamba}. While standard Mamba struggles with 2D visual data, the SS2D overcomes this through its Cross-Scan Module (CSM) (Fig.~\ref{fig:1}(e)), which scans image patches in four directions (top-left, bottom-right, top-right, bottom-left) to generate independent sequences. These sequences are processed through the Selective Scan Space State (S6) model \cite{gu2023mamba} before fusion, enabling comprehensive global receptive fields with minimal computational overhead. IceMamba employs patch embedding to convert inputs into numerical representations, dynamically capturing spatial relationships through CSM's ordered sequence transformation and S6's parallel cyclic processing, eliminating positional encoding requirements. Spatial resolution adjustments use patch merging/expanding instead of traditional pooling/upsampling: merging concatenates adjacent patches to halve resolution while doubling channels, whereas expanding splits patches to increase resolution, proportionally reducing channels (Sect.~\ref{sec4} details RESSB implementation). This design reduces computational complexity while enhancing generalization across visual perception tasks.

The encoder module analyzes input data, exploiting spatiotemporal and ice-atmosphere-ocean coupling relationships at different scales. This process produces a set of feature maps, each capturing a different aspect of the input sequences. After that, the decoder module gradually restores the scale of the feature maps, ultimately outputting the monthly SIC for the next several months. The feature maps captured by the encoder and decoder at the corresponding levels are fused by skip connection to integrate the spatial dependencies at these scales. The final patch-expanding layer increases the spatial resolution without reducing the number of channels. It ensures that the output tensor retains the full channel information, which is important since the channels encode temporal information. Following this layer, a 1×1 convolution is applied to generate the final output. Reducing the number of channels at this stage could result in the loss of crucial temporal features. Tanh activation is used in the 1×1 convolution layer. We find that using the Tanh function leads to better model performance compared to the no activation function and the Sigmoid activation function in prediction. We hypothesize that this is because the Tanh function provides a zero-centered distribution. This helps maintain gradient stability during training, preventing issues like vanishing and exploding gradients. Additionally, the Tanh function better captures non-linear features, leading to improved model performance when learning complex data patterns.

\bmhead{Residual efficient state space blocks}
We proposed the Residual Efficient State Space Blocks (RESSB) as a modification of the VSSB for sea ice forecast. The RESSB is formulated as:
\begin{equation}
\begin{split}
\mathbf{F}^{N+1}_{E} &= \mathrm{VSSB}(\mathrm{VSSB}(\mathrm{ECA}(\mathbf{F}^{N}_R))),\\
\mathbf{F}^{N+1}_{C} &=  \mathrm{SiLU}(\mathrm{Conv_{1\times1}} (\mathbf{F}^{N}_R)),\\
\mathbf{F}^{N+1}_{R} &= \mathbf{F}^{N+1}_{E} \oplus  \mathbf{F}^{N+1}_{C} ,
\end{split}
\end{equation}
where $\mathbf{F}^{N}_{R}$ represents the feature map of the n-layer. $\mathbf{F}^{N+1}_{E}$ and $\mathbf{F}^{N+1}_{C}$ are the output features of the two branches in RESSB respectively. $\mathbf{F}^{N+1}_{R}$ represents the final output feature of RESSB. VSSB($\cdot$), and ECA($\cdot$) denote the VSSB, and ECA operations as shown in Fig.~\ref{fig:1}(c) and Fig.~\ref{fig:1}(d), respectively. $\mathrm{Conv_{1\times1}}(\cdot)$ represents a convolutional layer with a kernel size of 1. The ECA module incorporates an efficient channel attention mechanism that starts with Global Average Pooling (GAP) to aggregate features across each channel without any dimensionality reduction, thereby preserving original correspondence between channels. Subsequently, the ECA module utilizes 1D convolution to capture the interactions between local cross-channels. Finally, The ECA module computes the weights for each channel in a parameter-efficient manner and then activates them using a Sigmoid function to obtain the final channel attention weights.

Given the interactions between sea ice, atmosphere, and ocean data at different time scales, we leverage the powerful modeling capabilities of VSSB to capture long-range dependencies. To this end, we integrate all SIC, ocean, and atmosphere data into the input channels of IceMamba. However, it's worth noting that while VSSB excels at retaining long-range dependencies, it requires the introduction of a large number of hidden states. This tendency often leads to apparent channel redundancy \cite{liu2024vmamba}. Hence, we incorporate ECA into RESSB, which allows the block to focus on learning diverse channel representations, while the channel attention mechanism selects the most critical channels to prevent redundancy. By integrating the ECA, we can alleviate this issue and improve the block's overall performance. Moreover, we incorporated a residual branch \cite{he2016deep} with a $1 \times 1$ convolutional layer to enhance the block's capability to capture complex patterns. The residual branch enables the module to learn residual information, while the $1 \times 1$ convolutional layer enables the block to better fuse the information from the residual branch.

\bmhead{Principles and application of the State space model for SIC forecast}

The state space model (SSM) is inspired by linear time-invariant systems. As shown in Fig.~\ref{fig:1} (e), the Cross-Scan Module (CSM) maps input data to four types of vectors, and each vector will be input into a corresponding Selective Scan Space State Sequential (S6) model. Assume that one of the vectors is $x(t) \in \mathbb{R}^L$, representing the input data to the S6 model, then an intermediate state $h(t) \in \mathbb{R}^N$ is used to map the input data to an output response $y(t) \in \mathbb{R}^L$, representing the output feature. This can be mathematically expressed using a linear ordinary differential equation (ODE):

\begin{equation}
\begin{split}
h'(t) &= \mathbf{A}h(t) + \mathbf{B}x(t), \\
y(t) &= \mathbf{C}h(t) + \mathbf{D}x(t),\label{eq:6}
\end{split}
\end{equation}
where $N$ is the state size, $\mathbf{A} \in \mathbb{C}^{N \times N}$ is the state transition matrix, $\mathbf{B} \in \mathbb{C}^{N}$ and $\mathbf{C} \in \mathbb{C}^{N}$ are the input and output projection parameters, and $\mathbf{D} \in \mathbb{C}^1$ is the skip connection.

For practical implementation in DL algorithms, the continuous-time SSMs can be discretized by the zero-order hold (ZOH) method, which allows for the conversion of continuous parameters $\mathbf{A}$ and $\mathbf{B}$ into discrete-time parameters $\mathbf{\overline{A}}$ and $\mathbf{\overline{B}}$:

\begin{equation}
\begin{split}
\mathbf{\overline{A}} &= e^{\Delta \mathbf{A}}, \\
\mathbf{\overline{B}} &= (e^{\Delta \mathbf{A}} - \mathbf{I})\mathbf{A^{-1}} \mathbf{B} \approx (\Delta \mathbf{A})(\Delta \mathbf{A})^{-1} \Delta \mathbf{B} = \Delta \mathbf{B},
\end{split}
\end{equation}
where $\mathbf{I}$ is the identity matrix and $\Delta$ is the timescale parameter to transform the continuous parameters $\mathbf{A}$ and $\mathbf{B}$ to their discrete counterparts $\mathbf{\overline{A}}$ and $\mathbf{\overline{B}}$. For the S6 model, time-variability and nonlinearity are introduced to the state-space model (SSM) by mapping the input data $\mathbf{x}$ to the parameter $\Delta$ through a nonlinear function. Specifically, $\mathbf{B}$ and $\mathbf{C}$ also depend on the input $\mathbf{x}$, generated dynamically via the nonlinear mappings. This design allows the S6 model to have time-varying system parameters at each time step, enabling the state-space model to handle both time-variability and nonlinear relationships.

The discretized version of Eq.~\ref{eq:6} can be rewritten in recurrent neural network (RNN) form as:

\begin{equation}
\begin{split}
h_k &= \mathbf{\overline{A}}h_{k-1} + \mathbf{\overline{B}}x_{k}, \\
y_k &= \mathbf{C}h_{k} + \mathbf{D}x_{k},
\end{split}
\end{equation}
where $k$ represents the initialization month for prediction. $x_k$ represents the input vector at the $k$-th initialization month, which includes the feature for SIC and reanalysis variables for the past 12 months leading up to the initialization month. The output $y_k$ represents the output feature for the predicted SIC for the $k$-th initialization month. The state $h_k$ captures the temporal dependencies and system dynamics up to the initialization month $k$, serving as the state at the initialization month for prediction. Moreover, Eq.~\ref{eq:6} can be transformed into a convolutional neural network (CNN) structure resulting in the following formulation \cite{gu2023mamba}:

\begin{equation}
\begin{split}
\mathbf{y} &= \mathbf{x} * \mathbf{\overline{K}}, \\
\mathbf{\overline{K}} &= [\mathbf{C}\mathbf{\overline{B}}, \mathbf{C}\mathbf{\overline{A}}\mathbf{\overline{B}}, \ldots, \mathbf{C}\mathbf{\overline{A}}^{L-1}\mathbf{\overline{B}}],
\end{split}
\end{equation}
where $\mathbf{\overline{K}} \in \mathbb{R}^L$ is a structured convolutional kernel and $*$ denotes the convolution operation. This CNN-based representation enables parallel training of recurrent-based models and allows the model to learn and capture the spatial and temporal patterns in the SIC and climate data effectively.

\bmhead{The detail of training sheme}
In the training phase, the network receives input from randomly selected batches of training data. The adaptive moment estimation (Adam) \cite{kingma2014adam} optimizer is employed to minimize the loss function. To ensure the model concentrates on sea ice within the pan-Arctic region, the loss function (MAE) is computed exclusively for non-land areas. A batch size of 1 is used with an initial learning rate of 0.001. To ensure the convergent and robust performance of the model, a learning rate schedule is defined for training. Specifically, the learning rate is decreased by 0.5 every 10 epochs, which helps the model avoid oscillations and overfitting in the later training stages. Additionally, an early stopping strategy of 10 epochs is implemented to prevent overfitting and enhance the model's generalization. This strategy allows the model to stop training when the validation loss stops improving, which helps to prevent the model from overfitting the training data. IceMamba is developed in Python 3.7, utilizing the PyTorch deep learning framework. The entire computational process is executed on an Nvidia A100 GPU, which enables the IceMamba training to be completed in approximately four hours.

\bmhead{Data availability}

The datasets used in this study include observational SIC data and climate reanalysis data, both accessible online. The SIC data is sourced from the NSIDC, available at \url{https://nsidc.org/data/g02202/versions/4}. The reanalysis data is derived from the ERA5 and ORAS5 datasets, with ERA5 single-level variables retrievable at \url{https://cds.climate.copernicus.eu/ cdsapp#!/dataset/reanalysis-era5-single-levels-monthly-means}, ERA5 pressure-level variables at \url{https://cds.climate.copernicus.eu/cdsapp#!/dataset/reanalysis-era5-pressurelevels-monthly-means}, ORAS5 variables at \url{https://cds.climate.copernicus.eu/datasets/reanalysis-oras5?tab=overview}.

\bmhead{Code availability}The source code used for the design, training, and evaluation of the IceMamba model, as well as the scripts for data preprocessing and analysis, are available on \url{https://github.com/WeiWang31/IceMamba.git}.

\bmhead{Acknowledgements}
This work is supported by the National Natural Science Foundation of China (grant No. 42325604).
The computations in this research were performed using the CFFF platform of Fudan University.

\bmhead{Author contributions}
Wei Wang designed and trained IceMamba, and wrote the manuscript under the guidance of Lei Wang and Weidong Yang. Wei Wang handled the climate data downloading and preprocessing under the supervision of Guihua Wang and Ruibo Lei. Wei Wang conducted the multi-model comparison with the assistance of Lei Wang. Weidong Yang managed the project. All authors reviewed and provided feedback on the manuscript.

\bmhead{Competing interests}The authors declare no competing interests.






\bibliography{sn-bibliography}

\end{document}